\documentclass[10pt,twocolumn,letterpaper]{article}

\usepackage{cvpr}
\usepackage{times}
\usepackage{epsfig}
\usepackage{graphicx}
\usepackage{amsmath}
\usepackage{amssymb}
\usepackage{multirow}
\usepackage{color}
\usepackage{subcaption}
\usepackage[breaklinks=true,bookmarks=false]{hyperref}

\cvprfinalcopy % *** Uncomment this line for the final submission

\def\cvprPaperID{****} % *** Enter the CVPR Paper ID here

\begin{document}

%%%%%%%%% TITLE
\title{DFS: A Diverse Feature Synthesis Model for Generalized Zero-Shot Learning }

\author{Bonan Li\\
University of Chinese Academy of Sciences\\
{\tt\small libonan16@mails.ucas.ac.cn}
\and
Xuecheng Nie \footnotemark[1]\\
Yitu Technology\\
{\tt\small xuecheng.nie@yitu-inc.com}
\and
Congying Han \footnotemark[2]\\
University of Chinese Academy of Sciences\\
{\tt\small hancy@ucas.ac.cn}
}

\maketitle
\footnotetext[1]{This work is supported by Yitu Technology}
\footnotetext[2]{Corresponding author}
%%%%%%%%% ABSTRACT
\begin{abstract}
Generative based strategy has shown great potential in the Generalized Zero-Shot Learning task. However, it suffers severe generalization problem due to lacking of feature diversity for unseen classes to train a good classifier. In this paper, we propose to enhance the generalizability of GZSL models via improving feature diversity of unseen classes. For this purpose, we present a novel Diverse Feature Synthesis (DFS) model. Different from prior works that solely utilize semantic knowledge in the generation process, DFS leverages visual knowledge with semantic one in a unified way, thus deriving class-specific diverse feature samples and leading to robust classifier for recognizing both seen and unseen classes in the testing phase. To simplify the learning, DFS represents visual and semantic knowledge in the aligned space, making it able to produce good feature samples with a low-complexity implementation. Accordingly, DFS is composed of two consecutive generators: an aligned feature generator, transferring semantic and visual representations into aligned features; a synthesized feature generator, producing diverse feature samples of unseen classes in the aligned space. 
We conduct comprehensive experiments to verify the efficacy of DFS. Results demonstrate its effectiveness to generate diverse features for unseen classes, leading to superior performance on multiple benchmarks. Code will be released upon acceptance.
\end{abstract}
\section{Introduction}

Generalized Zero-Shot Learning (GZSL) is an important yet challenging problem in computer vision, aiming to recognize object categories unseen in the training phase. It is widely utilized in various applications, \eg, image recognition~\cite{luo2020context,kumar2020harnessing,Brattoli_2020_CVPR,zhan2019zero}, object detection~\cite{rahman2020improved,zhou2020motion,li2020consistent} and super-resolution~\cite{Soh_2020_CVPR}, due to the powerfulness of facilitating a model to reason class information of new objects with only their semantic correlations to known ones. 
\begin{figure}[t!]
\begin{center}
  \includegraphics[width=0.5\textwidth]{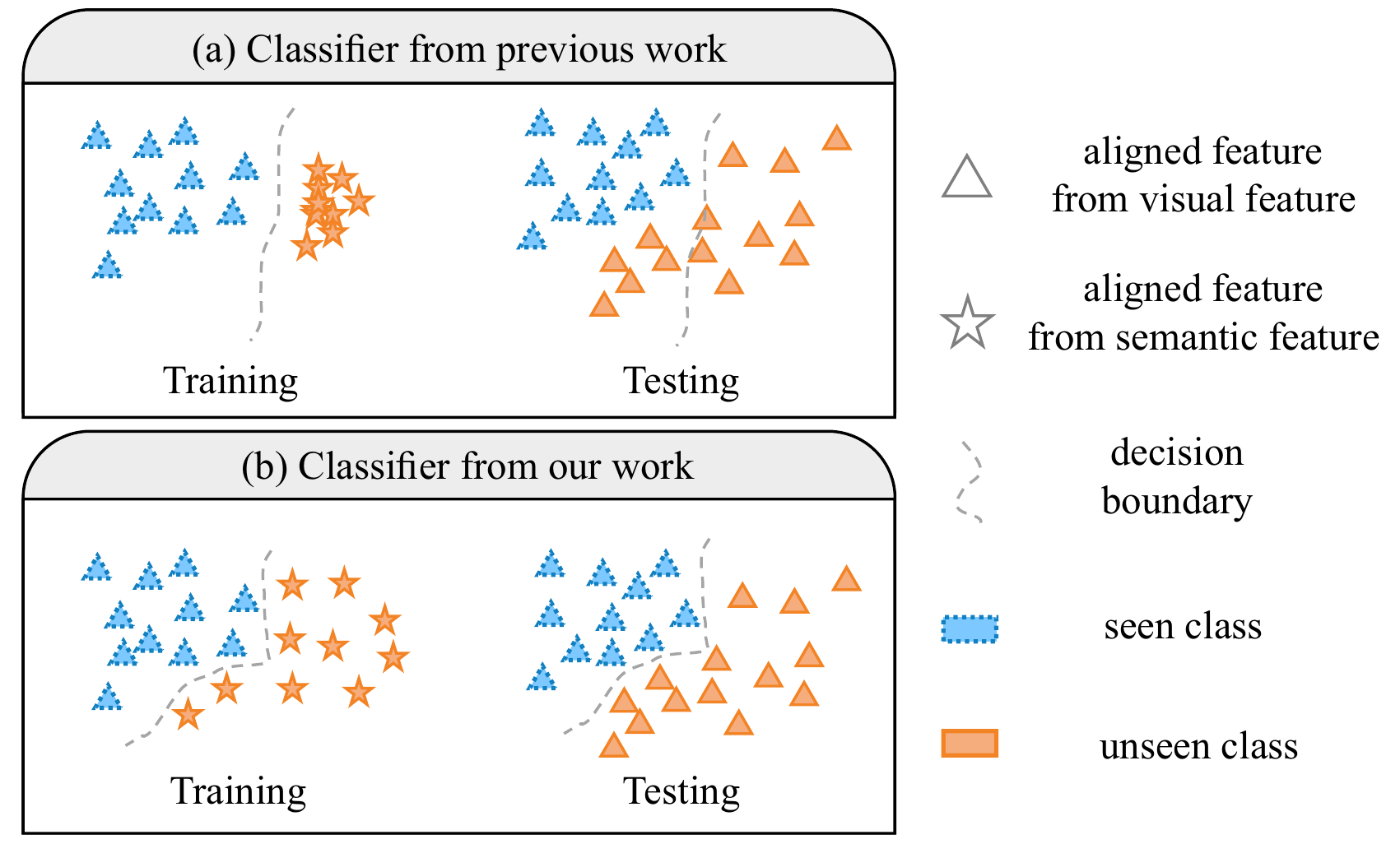}
\end{center}
  \vspace{-3mm}
  \caption{Motivation of our Diverse Feature Synthesis model for the Generalized Zero-Shot Learning task. (a) Prior works fail to generate diverse features of unseen classes, causing classifier trained with these samples poorly generalize to testing scenarios. (b) Our DFS model can enhance feature diversity for unseen classes, thus improving the generalizability of classifiers for the GZSL task.}
\label{fig:classifier}
\end{figure}

Existing methods to tackle the GZSL task mainly follow two strategies: one is embedding based~\cite{lampert2013attribute,norouzi2013zero,socher2013zero,zhang2015zero,changpinyo2016synthesized,kodirov2017semantic}, that learns an embedding model between visual and semantic spaces, following with nearest neighbor search for deriving class cues; the other is generation based~\cite{xian2018feature,gao2020zero,narayan2020latent,hubert2017learning,schonfeld2019generalized,mavariational}, that applies generative model to produce features for unseen classes, converting the problem to a traditional classification problem. Recently, generative based strategy has shown dominated performance over embedding based one, regarding to its capability of alleviating the seen-bias problem. Despite its successful, generative based strategy still faces two major problems, that heavily hurt its generalizability while are ignored by prior works. The first problem is lack of diversity for generated features of unseen classes, leading to inaccurate classification boundary. As shown in Figure~\ref{fig:classifier}(a), generated samples from prior works are platitudinous and cannot describe the true distribution of unseen classes, causing classifier trained with these samples performs poorly in the testing phase. The second problem is high-complexity of feature generator, due to employing GAN with complex settings, which results in unstable training procedure~\cite{xian2018feature,zhu2018generative,huang2019generative,narayan2020latent}. Motivated by this, we propose to boost the generalizability of GZSL models via enhancing the feature diversity with low-complexity generators.

To achieve this goal, we present a novel Diverse Feature Synthesis (DFS) model for the GZSL task in this paper. 
Generally, semantic knowledge has specificness, \eg, one class label only maps to a particular semantic feature; Whereas, visual knowledge has diversity, \eg, one class label maps to multiple visual features. Prior works~\cite{ hubert2017learning,schonfeld2019generalized,mavariational} generate feature samples for unseen class solely relied on their semantic knowledge. Differently, our DFS model proposes to incorporate visual and semantic knowledge, together, for inheriting both of their properties. Thus, DFS can generate diverse features for a specific unseen class. Besides, instead of visual space, DFS performs feature generation and category classification in the aligned space due to its simplicity and descriptiveness, therefore, alleviating the stress for capturing feature diversity and making a low-complexity model competent for the task. 

In particular, DFS is implemented with two generators:
%we implement DFS with two moudles, 
an Aligned Feature Generator (AFG) and a Synthetic Feature Generator (SFG). DFS first utilizes AFG to transfer features from semantic and visual spaces into an aligned space, which is derived from the cross model learning~\cite{schonfeld2019generalized}.  
%AFG is used to project semantic and visual features into an aligned space. In this work, we implement AFG based on powerful cross model~\cite{schonfeld2019generalized}. 
Then, DFS learns SFG, designed as a Conditional Variational AutoEncoder (CVAE), with two steps: In the first step, SFG feeds aligned semantic and visual features of \emph{seen classes} as input to the encoder, which produces latent features embedding with both semantic and visual knowledge; In the second step, SFG feeds the latent features into its decoder to reconstruct aligned visual features of \emph{seen classes}, given the semantic knowledge as condition. After SFG learned, in the inference phase, DFS removes its encoder and only uses the decoder to generate feature samples for \emph{unseen classes}. Here, latent features for unseen classes are synthesized via randomly sampling from standard Gaussian distribution. 
%SFG takes the semantic and visual representation derived from the AFG as input to its encoder, which produces latent features embedding with both semantic and visual knowledge. Then, SFG feeds the latent features into its decoder to reconstruct the visual knowledge, given the semantic knowledge as condition. Then, SFG produces the latent feature via randomly sampling from standard Gaussian distribution. Finally, SFG inputs the aligned semantic representation and latent features into its decoder to generate visual representation for training the classifier of ZSL model. 
In this way, DFS introduces visual knowledge into the feature generation process, it significantly increases the diversity of generated samples, overcoming the drawbacks of prior works and leading to improved classifier with better generalizability, as shown in Figure~\ref{fig:classifier}(b). In addition, usage of aligned features simplifies the feature generation process, enabling a low-complexity VAE to satisfy the requirement.
%all modules in DFS are based on VAE components, so a low-complexity model is constructed. 
The overall framework of DFS is shown in Figure~\ref{fig:model}.

Comprehensive experiments verify the efficacy of DFS to generate diverse feature samples for unseen classes as well as to improve  
%generated from DFS as well as verify its effectiveness for improving 
the generalizability of GZSL models.
%on multiple benchmarks. 
Our contributions are in two folds: (1) We propose a novel model for effectively and efficiently generating diverse features for unseen classes for the GZSL task; 
%(2) We presents a low-complexity implementation for feature generation
%we propose a low-complexity model for simplifying the feature generation process; 
(2) With feature samples generated from our model, we set new state-of-the-arts on multiple benchmarks for the GZSL task.

% We conduct comprehensive experiments to evaluate the diversity of feature samples generated from the proposed DFS and also verify its effectiveness for improving the generalizability of ZSL models on multiple benchmarks. Our contributions are summarised into three folds: (1) we propose an effective way for generating diverse aligned features for unseen classes through incorporating visual and semantic knowledge; (2) we propose a low-complexity model for simplifying the feature generation process; (3) with feature samples generated from our model, we set new state-of-the-arts on multiple benchmarks for the GZSL task.

\section{Related Work}
\label{related}

% First introduce what is ZSL, CZSL and GZSL.
In literature, Zero-Shot Learning (ZSL) has been well studied. 
%in recent years. 
It can be categorized into Conventional ZSL (CZSL) and Generalized ZSL (GZSL) depending on the classes contained in the testing dataset.
For CZSL, the testing dataset only contains unseen classes samples. 
However, for GZSL, both seen and unseen classes samples are included in the testing dataset.
Compared with CZSL, GZSL is more practical, and most of current researches in ZSL area aim at solving this problem.

% Then introduce embedding based methods.
Early ZSL approaches are mainly based on embedded models and can be divided into 
%the following 
three groups. 
%according to the embedded space. 
The first group ~\cite{lampert2013attribute,romera2015embarrassingly} learns a projection function from visual feature space to a semantic space.
%Traditional regression or ranking models~\cite{akata2013label,akata2015evaluation} and deep neural network regression or ranking models ~\cite{lei2015predicting,frome2013devise,reed2016learning,socher2013zero} are widely used in these methods. 
The second group ~\cite{zhang2017learning,annadani2018preserving} of approach map semantic features to visual space. 
%In general, higher dimensional feature spaces tend to alleviate the hubness problem~\cite{radovanovic2010hubs} caused by nearest neighbor searches.  
The third group ~\cite{changpinyo2016synthesized,romera2015embarrassingly,zhang2016zero,hubert2017learning} adopts latent space to establish mapping between semantic and visual domains. % The mapping function directly established between the features of distinct domain is often unstable. Thus, the common latent space is adopted by final group as the embedding space
Although the above methods have achieved remarkable results in CZSL setting, these models will produce obvious bias to the visible classes in GZSL setting.
This is demonstrated by the fact that the classification accuracy of seen classes is much higher than that of unseen classes.

% After that, introduce generative based methods. In this section, point out the difference of our method and prior works.
Recently, the powerful generative methods, \eg Generative Adversarial Network(GAN)~\cite{goodfellow2014generative} and Variational Autoencoder(VAE)~\cite{kingma2013auto}, are utilized to synthesize massive features of unseen classes from prototype vector~\cite{verma2017simple,xian2018feature,kumar2018generalized,paul2019semantically,keshari2020generalized,mishra2018generative,gao2020zero,zhu2018generative}.
%In the final stage, t
These synthesized features of unseen classes will be used together with features of seen classes to train a fully supervised classifier. 
This way 
%The introduction of unseen class features 
can promote generalization of the classifier, thus reducing the bias to seen classes, resulting in a higher harmonic mean.
%but the high complexity of the target space and the scarcity of data samples prevent the network from accurately capturing the real distribution in visual space
f-CLSWGAN~\cite{xian2018feature} applies GAN to generate visual features conditioned on semantic features, but it suffers from mode collapse issues and unstable training phase~\cite{arjovsky2017towards}. 
%On the other hand, although 
VAE based algorithm~\cite{mishra2018generative,keshari2020generalized} can train stably, but it fails to capture the complex distribution~\cite{bao2017cvae}, leading to unsatisfied results. 
%Thus, the results achieved by are not satisfactory.
In order to overcome the above shortcomings, \cite{xian2019f,gao2020zero} combines
two generative models, \ie, VAE/GAN, to generate samples for unseen classes. %TF-VAEGAN~\cite{narayan2020latent} utilizes the semantic embedding decoder at all stages to exploit complementary information with respect to samples. 
Despite of achieving performance improvement,
%While the introduction of these additional modules improves performance, 
their complex parameter setting and tedious training process can not be ignored.
In contrast to the above methods, the works~\cite{schonfeld2019generalized,mavariational} train the cross model to encode and decode features from semantic and visual modalities by matching their parameterized distributions and joining a cross-modal reconstruction term. Although ~\cite{schonfeld2019generalized,mavariational} exhibits superior results with stable training, 
%we argue that 
they are still unable to effectively generate massive significantly different features with the same semantic information, which would lead to the classifier inevitably biasing seen classes. To alleviate these two problems at the same time, we take a stable model to synthesize more diverse features of unseen classes. Details are depicted in next section.

% Later works~\cite{gao2020zero,felix2018multi,li2019leveraging,jiang2019transferable,narayan2020latent} attempts to solve this problems through a variety of generative models and regularization terms. Although some good results have been achieved, complex parameters and unstable training methods have been introduced.
% In order to build a stable training model, Schonfeld \etal~\cite{schonfeld2019generalized} train a cross model to encode and decode features from semantic and visual modalities by matching their parameterized distributions and joining a cross-modal reconstruction term. 
% Then it make one class embedding can generate several samples in aligned space by multi-sampling from mean and variance.
% Although CADA-VAE~\cite{schonfeld2019generalized} exhibits superior results, we argue that it is still unable to effectively generate massive significantly different features with the same semantic information, which would lead to the classifier inevitably biasing seen classes.
% To overcome this problem, in aligned space, we take an extra stable module to synthesize more diverse features of unseen classes to help improve the generalizability of the classifier.

%-------------------------------------------------------------------------
\section{Method}
\label{method}
\subsection{Problem Definition}
We first depict the mathematical formulation for the Generalized Zero-Shot Learning (GZSL) problem. Let  $\mathcal{S}=\{(v_s,a_s,l_s)|v\in{V}_{s},a_s\in{A}_{s},l_s\in{L}_{s}\}$ denote the training set for seen classes, where $v_s$ is the visual feature of an image, $l_s$ is the corresponding class label, and $a_s$ is the semantic embedding for class $l_s$. Let $\mathcal{U}=\{(v_u,a_u,l_u)| v_u\in{V}_{u},a_u\in{A}_{u}, l_u\in{L}_{u}\}$ denote the testing set for unseen classes, where $v_u$, $a_u$, and $l_u$ are similarly defined as  $v_s$, $a_s$, and $l_s$, but ${L}_{s}\cap{L}_{u}=\emptyset$, meaning that seen classes and unseen ones are disjoint. Given $\mathcal{S}$, $A_u$ and $L_u$, GZSL targets at learning a function $f$ that can recognise both seen and unseen classes, 
% \emph{i.e.}
% \begin{equation}
%     f: v \rightarrow l,
% \end{equation}

$$f: v \rightarrow l$$
where $v \in V_s \cup V_u$ and $l \in L_s \cup L_u$.

To solve the GZSL task, generative based strategy models $f$ as a classifier through converting the original problem to the traditional classification problem. Accordingly, its core is to generate visual features $\hat{V_u}$ for unseen classes, thus forming the training set $\mathcal{T} = \{(v, l) | v \in V_s \cup \hat{V_u}, l \in L_s \cup L_u\}$ to learn the classifier $f$. For getting $\hat{V_u}$, prior works mainly follow two ways: (1) performing feature generation in the \emph{original} visual space with a Generative Adversarial Network based generator
%utilize a generator 
$g_{1}$, formulated as 
$$g_1: A_u, z \rightarrow \hat{V_u}$$
where 
%$g_{1}$ is often implemented with a Generative Adversarial Network (GAN), 
$z$ is a noise sampled from standard Gaussian distribution; (2) performing feature generation in the \emph{aligned} space with a sampler $g_2$:
$$g_2: H(A_u) \rightarrow H(\hat{V_u})$$
where $H(\cdot)$ is a function for transferring features from original space to aligned space.
%the aligned feature generator, $g_2$ is implemented with the sampling method.
However, the first way is always built on a complex model to mitigate the large gap between semantic and visual knowledge, and the second way often suffers from samples lacking of diversity. 

Differently, in this paper, we propose a Diverse Feature Synthesis (DFS) model, defined by $\hat{g}$, to generate diverse feature samples for unseen classes in the aligned space, %$H(\hat{V_u})$ , 
as
% \begin{equation}
%     \hat{g}: H(A_u), z \rightarrow H(\hat{V_u}).
% \end{equation}
$$\hat{g}: H(A_u), z \rightarrow H(\hat{V_u})$$
In this way, DFS is able to effectively generate diverse feature samples for unseen class via low-complexity model.
% leveraging visual knowledge. 
Thus, DFS overcomes drawbacks of previous and leads to a more accurate classifier $f$ for the GZSL task. In next subsection, we will illustrate the implementation details of the proposed DFS model.

\begin{figure*}[t!]
\begin{center}
  \includegraphics[width=0.95\textwidth]{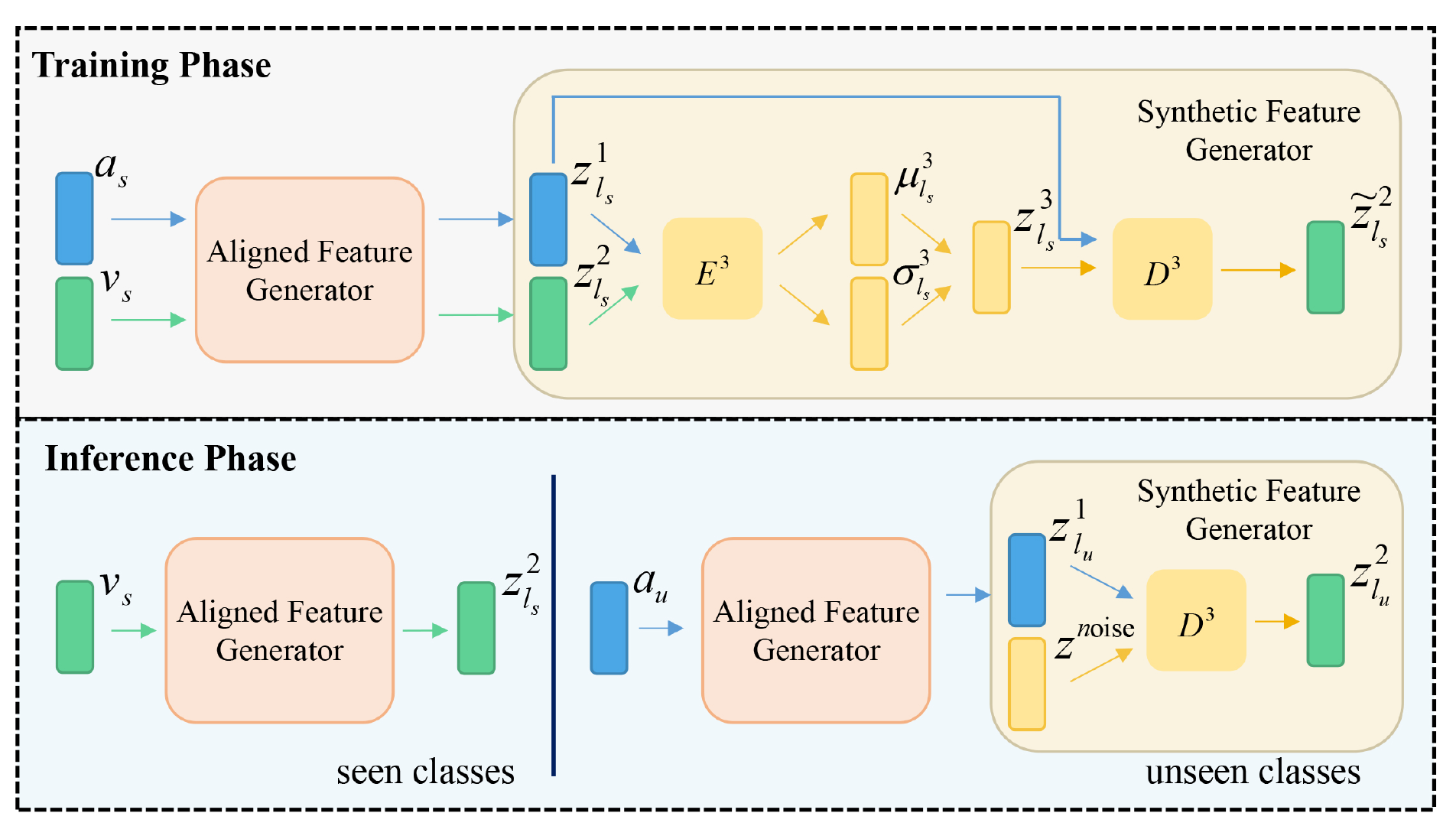}
\end{center}
 \caption{The overview of our proposed Diverse Feature Synthesis (DFS) model for the Generalized Zero-Shot Learning (GZSL) tasks. Top: the \emph{training} phase. DFS first utilizes an aligned feature generator to transfer semantic and visual features of seen classes, $a_s$ and $v_s$ into aligned space, \ie, $z_{l_s}^1$ and $z_{l_s}^2$. Then, it employs a synthesised feature generator to produce synthesised features. Specifically, it feeds the aligned features into the decoder $E^3$ to generate the latent feature $z_{l_s}^3$ via sampling from $\mu_{l_s}^3$ and $\sigma_{l_s}^3$ and reconstructs the visual representation by the decoder $D^3$ with $\tilde{z}_{l_s}^2$. Bottom: the \emph{inference} phase. For seen classes, the aligned feature $z_{l_s}^2$ are directly produced by the aligned feature generator, given $v_s$ as input. While for unseen classes, DFS removes the encoder of the synthesised feature generator. After deriving aligned feature $z_{l_u}^1$ from $a_u$, it synthesises the latent feature $z^{noise}$ via random sampling from standard Gaussian distribution for input to $D^3$ to generate the feature sample $z_{l_u}^2$ of unseen class.}
  %Illustration of the proposed model for synthetising diverse features in aligned space. 
  %In training phase, the pre-trained aligned feature generator is used to obtain the aligned feature $z^{1}_{l_s}$ and $z^{2}_{l_s}$ of semantic and visual features in aligned space, respectively. Subsequently, the framework uses $z^{1}_{l_s}$ and $z^{2}_{l_s}$ as input to train synthetic feature generator, $E^{3}$ and $D^{3}$. In inference phase, the instances of seen class are directly from aligned feature generator. For unseen class, the instances are generated by DFS with $z^{1}_{l_u}$ and $z^{noise}$ where $z^{noise}\sim \mathcal{N}(0,1)$. 
  
\label{fig:model}
\end{figure*}

\subsection{Diverse Feature Synthesis Model}
In this section, we will explain the implementation of our DFS model in details, including its network architecture as well as the training and inference phases. 

\subsubsection{Network Architecture}
Our DFS model is composed of two modules: an Aligned Feature Generator and a Synthetic Feature Generator. Their details will be illustrated in the following, respectively.

\paragraph{Aligned Feature Generator}
\label{sec:first_stage}
We implement the aligned feature generator based on the recently proposed model CADA-VAE~\cite{schonfeld2019generalized}, which achieves impressive results for generalized zero-shot learning by using a stable cross VAE model.
Nevertheless, we are not limited to choosing CADA-VAE model as the basis and our DFS network can still be effective in improving performance when other aligned feature generators are selected as the baseline.
More concretely, two encoders, $E^{1}$ and $E^{2}$, first encode semantic and visual vectors as aligned features, $z^{1}_{l_s}$ and $z^{2}_{l_s}$, respectively.
After obtaining the aligned features, we take them passed through the decoders, $D^{1}$and $D^{2}$, to generate reconstruction features which have the same dimensions with original input vectors.
According to classical VAE model, we can formulate the loss as follows:
\begin{equation}
\label{eq:VAE_l}
\begin{split}
\mathcal{L}_{VAE}&= \sum^{M}_{k=1}\mathbb{E}_{q_{\phi}(z^{k}_{l_s}|x^{k}_{l_s})[\log p_{\theta}(x^{k}_{l_s}|z^{k}_{l_s})]}\\
&- \beta_{1} D_{KL}(q_{\phi}(z^{k}_{l_s}|x^{k}_{l_s})||p_{\theta}(z^{k}_{l_s}))
\end{split}
\end{equation}
where $q_{\phi}(z^{k}_{l_s}|x^{k}_{l_s})$ is modeled as $E^{k}$, $p_{\theta}(z^{k}_{l_s})$ is assumed to be $\mathcal{N}(0,1)$,  $p_{\theta}(x^{k}_{l_s}|z^{k}_{l_s})$ is equal to $D^{k}$ and $D_{KL}(\cdot)$ denotes KL-Divergence. $\beta_{1}$ is the hyper-parameter to weight the loss of KL-Divergence and restruction loss. 
$z$ is the feature in aligned space. For ZSL, the features are usually draw from visual and semantic domain, so we set $M=2$, $x^{1}_{l_s} \in A_{s}$ and $x^{2}_{l_s} \in V_{s}$.

Here, in order to learn representations within an aligned  space, two extra loss terms which named Distribution-Alignment loss ($\mathcal{L}_{DA}$) and Cross-Reconstruction loss ($\mathcal{L}_{CA}$) are introduced into model .

$\mathcal{L}_{DA}$ is mainly used to minimize the Wasserstein distance between the latent multivariate Gaussian distributions to ensure the consistency of different modalities. The specific forms are as follows:
\begin{equation}
\label{equ:DA_L}
\mathcal{L}_{DA}=\sum^{M}_{k=1}\sum^{M}_{t=1 \atop t\ne k}(||\mu^{k}_{l_s}-\mu^{t}_{l_s}||^{2}_{2}+||(\sigma^{k}_{l_s})^{\frac{1}{2}}-(\sigma^{t}_{l_s})^{\frac{1}{2}}||^{2}_{Frobenius})^{\frac{1}{2}}
\end{equation}
where $\mu^{k}_{l_s}$ is mean and $\sigma^{k}_{l_s}$ is variance parameters for multivariate Gaussian distributions from $k^{th}$ modality.

Due to the aligned space contains the domain shared and transferable information, the modality-specific features should also can be reconstructed with the aligned feature of the same sample but from distinct modalities.
Thus, $\mathcal{L}_{CA}$ is defined as follows:
\begin{equation}
\label{eq:CA_L}
\mathcal{L}_{CA}=\sum^{M}_{k=1}\sum^{M}_{t=1\atop t\neq k}Dis(x^{k}_{l_s},D^{k}_{l_s}(z^{t}_{l_s}))
\end{equation}
where $z^{k}_{l_s}$ is the latent feature from $k^{th}$ modality and $z^{k}_{l_s}=E^{k}(x^{k}_{l_s})$. $Dis(\cdot)$ is the Manhattan distance function.

Thus at the end of this part, we encode the features of different modalities to an aligned space.
However, the model lack the ability to generate significantly diverse features for unseen classes.
\paragraph{Synthetic Feature Generator}
For one class $l\in L_s \cup L_u$, aligned embeddings from visual space usually sampled from multiple distributions, that is, $z^{2}_{l}$ can be sampled from $\{\mathcal{N}(\mu^{2}_{l,1},\sigma^{2}_{l,1}),\mathcal{N}(\mu^{2}_{l,2},\sigma^{2}_{l,2}),...,\mathcal{N}(\mu^{2}_{l,r},\sigma^{2}_{l,r})\}$, where $r$ denotes the $r^{th}$ sample in class $l$.
Nevertheless, aligned embeddings $z^{1}_{l}$ from semantic space only can be sampled from unitary distributions $\mathcal{N}(\mu^{1}_{l},\sigma^{1}_{l})$.
Since the variance of specific information in visual space is difficult to be captured directly by semantic features, these information is contained in $\mu^{2}_{l,r}$ as the unique feature of each sample.

In order to make the instances generated from semantic feature more diverse, in this part, a module named SFG is designed to explicitly capture the distribution of visual-specific information in aligned space. The framework of SFG is shown in Figure~\ref{fig:model}.
By given the paired features $(z^{1}_{l_s},z^{2}_{l_s})$ encoded in aligned space, an encoder $E^{3}$ is used to compute the latent parameters $\mu^{3}_{l_s}$ and $\sigma^{3}_{l_s}$.
After that, the feature $z^{3}_{l_s}$ is obtained by sampling from $\mathcal{N}(\mu^{3}_{l_s},\sigma^{3}_{l_s})$ with reparameterization trick.
The decoder $D^{3}$ reconstructs $z^{2}_{l_s}$ with $z^{3}_{l_s}$ and $z^{1}_{l_s}$ as input.
Therefore, both $E^{3}$ and $D^{3}$ are conditioned on the latent feature $z^{1}_{l_s}$ and then we can learn it with the follow loss:
\begin{equation}
\label{eq:CVAE}
\begin{split}
\mathcal{L}_{CVAE}&=\sum^{L_{s}}_{l_s=1}\mathbb{E}_{E^{3}(z^{1}_{l_s},z^{2}_{l_s})}[\log D^{3}(z^{3}_{l_s},z^{1}_{l_s})]\\
& -\beta_{2} D_{KL}(E^{3}(z^{1}_{l_s},z^{2}_{l_s})||p(z^{3}_{l_s}|z^{1}_{l_s}))
\end{split}
\end{equation}
where $D_{KL}(\cdot)$ is also denote KL-Divergence like Equation~\ref{eq:VAE_l}, $p(z^{3}_{l_s}|z^{1}_{l_s})$ is a prior distribution assumed to be $\mathcal{N}(0,1)$ and $\log D^{3}(z^{3}_{l_s},z^{1}_{l_s})$ is the loss of the reconstruction. $\beta_{2}$ is the hyper-parameter to weight the loss of these two items.
In contrast to~\cite{mishra2018generative}, using $z^{1}_{l_s}$ instead of original semantic feature as condition can benefit from the following two points. 
First, the learning process of the network would not be disturbed by domain-specific information in the semantic space.
Second, $z^{1}_{l_s}$ and target distribution are on the same manifold, which will further reduce the training difficulty of the network.
In particular, we let $z^{1}_{l_s} = \mu^{1}_{l_s}$ to stabilize the training process of the model.

After training SFG, We can provide diverse samples in aligned space for each unseen classes with the condition from aligned semantic embedding.

\subsubsection{Training and Inference}
\paragraph{Training}In training stage, we firstly learn the encoder ($E^{1},E^{2}$) and decoder ($D^{1},D^{2}$) of different modalities simultaneously by minimizing the combination of the three loss function terms. 
The objtctive can be formulated as follows:
\begin{equation}
\label{equ:first_stage}
\mathcal{L}_{AFG}= \mathcal{L}_{VAE} + \eta\mathcal{L}_{DA} + \delta\mathcal{L}_{CA}
\end{equation}
where $\eta$ and $\delta$ are the penalty regularization coefficients for the loss of two regularization, respectively.

Then, the parameters of alignment feature generator are fixed and only the parameters of $E^{3} $ and $D^{3}$ are optimized by minimizing: 
\begin{equation}
\label{eq:CVAE}
\mathcal{L}_{SFG} = \mathcal{L}_{CVAE}
\end{equation}

\paragraph{Inference}For each seen class $l_{s}\in{L}_{s}$, we generate the instances in the aligned space by sampling from $\{\mathcal{N}(\mu^{2}_{l_s,1},\sigma^{2}_{l_s,1}),\mathcal{N}(\mu^{2}_{l_s,2},\sigma^{2}_{l_s,2}),...,\mathcal{N}(\mu^{2}_{l_s,r},\sigma^{2}_{l_s,r})\}$ where $\mu^{2}_{l_s,r},\sigma^{2}_{l_s,r}$ are computed by $E^{2}$ with $r^{th}$ visual feature of class $l_s$ as input.
For each class $l_{u}\in{L}_{u}$, we first obtain the conditional feature $z^{1}_{l_u} = \mu^{1}_{l_u}$ by $E^{1}$ with the semantic feature of class $l_u$. 
Subsequently, we take a set of noises from Gaussian noises $\mathcal{N}(0,1)$ with the same dimension as $z^{3}$ and connect them with $z^{1}_{l_u}$ respectively. These connected features are input to $D^{3}$, thus generating multiple diverse samples for $l_u$ class.
The specific process can be seen the inference phase in Figure~\ref{fig:model}.

\section{Experiments}
\label{exe}
\begin{table}
\caption{Illustration on datasets used in our experiments
%Statistics of five datasets: AWA2, CUB, SUN, FLO and APY.
}
\label{Tab:Dataset}
\centering
\resizebox{0.45\textwidth}{!}{
\begin{tabular}{c|c|c|c|c|c}
\hline
\textbf{Dataset}& \textbf{Detail} & \textbf{Seen/Unseen Classes} & \textbf{Images} & \textbf{Visaul} & \textbf{Att}\\

\hline
AWA2 & coarse & 40/10 & 37322 & 2048 & 85 \\
CUB & fine & 150/50 & 11788 & 2048 & 312 \\
SUN & fine & 645/72 & 14340 & 2048 & 102 \\
FLO & fine & 82/20 & 8189 & 2048 & 1024 \\
APY & coarse & 20/12 & 15339 & 2048 & 64 \\
\hline
\end{tabular}}
\end{table}
\subsection{Experiment setup}
\paragraph{Datasets}We evaluate our framework on five benchmarking datasets: AWA2~\cite{xian2018zero}, CUB~\cite{welinder2010caltech}, SUN~\cite{patterson2012sun}, FLO~\cite{reed2016learning} and APY~\cite{farhadi2009describing}. They contain 50, 200, 717, 102 and 32 categories, respectively. Other details on these five datasets are listed in Table~\ref{Tab:Dataset} for reference.

% \textbf{AWA2}~\cite{xian2018zero} \emph{Animal with Attributes2} (AWA2) is the dataset introduced by Xian et al.~\cite{xian2018zero}  for that the images in AWA1~\cite{lampert2013attribute} are not publicly available. It comprises of 37,322 images belonging to 50 classes and also provides 85 continuous attribute values for each class. It has 40 seen training classes and 10 unseen testing classes.

% \textbf{CUB}~\cite{welinder2010caltech} The second is a fine-grained dataset which contains 11,788 images of 200 different bird species. In \emph{CaltechUCSD-Birds 200-2011} (CUB), each sample is annotated with 312 attributes. It has 150 seen classes for training and 50 unseen classes for testing.

% \textbf{SUN}~\cite{patterson2012sun} SUN consists of 14,340 images from 717 different scenes with class which labelled by 102 attribute vector. We use 645 classes for training and remaining 72 classes for testing.

% \textbf{FLO}~\cite{reed2016learning} FLO consists of 8,189 images from 102 different varieties of flowers with class which labelled by 1024-d embeddings from a character-based CNN-RNN. We use 82 classes for training and remaining 20 classes for testing.

% \textbf{APY}~\cite{farhadi2009describing} \emph{Attribute Pascal and Yahoo} (APY) is a coarse-grained dataset with 64 attributes. It has 15,339 pictures of 20 Pascal classes and 12 Yahoo classes. We use 20 classes for training and remaining 12 classes for testing

\paragraph{Visual Space and Dataset Split}The visual features with 2048 dimensions we use in all experiments are extracted by powerful deep Convolutional Neural Networks (CNN), ResNet~\cite{he2016deep}, which is pre-trained with ImageNet~\cite{russakovsky2015imagenet}. 
% Xian \etal~\cite{xian2018zero} present a novel Proposed Splitting (PS) method to ensure that the test class of any dataset never appears in training set of the pre-trained model. Thus, PS mitigates the problem that Standard Splitting would violate the zero-shot learning condition. 
In this work, we apply Proposed Splitting (PS) proposed by Xian \etal~\cite{xian2018zero} to all datasets.

\paragraph{Training Details}For aligned feature generator, we set the aligned space dimension of coarse-grained dataset (AWA2, APY) to 64 and fine-grained datasets (CUB, SUN, FLO) to 256, because fine-grained datasets often need more information to train an effective classifier.
For coarse-grained datasets, all the network setting comes from ~\cite{schonfeld2019generalized}.
Since a higher aligned space dimension is set for fine-grained datasets, we appropriately increase the dimensions of encoder and decoder.
Specifically, 6240 hidden units are used for $E^{1}$ and 4980 hidden units are used for $D^{1}$. The $E^{2}$ and $D^{2}$ for semantic domain have 3600 and 1330 hidden units, respectively.
In addition, all the hyper-parameters in this part are also follow the settings in ~\cite{schonfeld2019generalized}.
For synthetic feature generator, $E^{3}$ and $D^{3}$ network are set with one hidden layer and have 1990 and 1560 hidden units respectively.
$\beta_{2}$ is set to 0.6. 
The dimension of SFG can be fine tuned by the accuracy on the validation dataset, but it is worth noting that the training data of our final model come from the training dataset and validation dataset.
Learning rate is set to 0.00015 and training epoch is 100 across all the datasets.
After training, a linear classifier is used to classify in aligned space.

\paragraph{Evaluation Metric}We average the classification accuracy of each test class and report the average top accuracy as following:

\begin{equation}
\label{equ:Accu}
Acc_{u}=\frac{1}{|\mathcal U|}\sum_{{u}\in \mathcal U}\frac{acc_{u}}{|{u}|}
\end{equation}
where $acc_{u}$ denotes the number of correctly classified samples for unseen classes $l_u$.

In the GZSL setting, we report the harmonic mean of the accuracy over seen and unseen classes which is defined as:
\begin{equation}
\label{equ:H}
Acc_{H}=\frac{2*Acc_{s}*Acc_{u}}{Acc_{s}+Acc_{u}}
\end{equation}
where $Acc_{s}$ denotes the mean class accuracy on seen classes, and $Acc_{u}$ indicates the mean class accuracy on unseen class.

\begin{table*}[ht]
\caption{Comparison between our DFS model with SOTAs using the generalized zero-shot learnig settings. Here, $Acc_s$ is TOP-1 accuracy on seen data, $Acc_u$ is TOP-1 accuracy on unseen data, and $H$ is Harmonic Mean. Baseline reports the results of aligned feature generator. $\wr$ denotes methods based on embedded model, $\dagger$ denotes the methods based on GAN and $\ddagger$ denotes the methods based on VAE, respectively. The best results are in red, and the second best results are in blue. (Best viewed in colour)}
\label{Tab:GZSLPer}
\centering
\resizebox{\textwidth}{!}{
\begin{tabular}{c|c|ccc|ccc|ccc|ccc|ccc}
\hline
& \multirow{2}{*}{\textbf{Method}} & \multicolumn{3}{|c|}{\textbf{AWA2}} & \multicolumn{3}{|c|}{\textbf{CUB}} & \multicolumn{3}{|c|}{\textbf{SUN}}& \multicolumn{3}{|c|}{\textbf{FLO}} & \multicolumn{3}{|c}{\textbf{APY}}\\

& & $Acc_{s}$ & $Acc_{u}$ & $Acc_{H}$ & $Acc_{s}$ & $Acc_{u}$ & $Acc_{H}$& $Acc_{s}$ & $Acc_{u}$ & $Acc_{H}$ & $Acc_{s}$ & $Acc_{u}$ & $Acc_{H}$& $Acc_{s}$ & $Acc_{u}$ & $Acc_{H}$\\
\hline
\multirow{9}{*}{$\wr$} & DAP~\cite{lampert2013attribute} &  84.7 & 0.0 & 0.0 &  67.9 & 1.7 & 3.3 &  25.1 & 4.2 &  7.2 &- & - & - &  78.3 &  4.8 &  9.0\\
& IAP~\cite{lampert2013attribute} &  87.6 & 0.9 & 1.8 &  72.8 & 0.2 & 0.4 &  37.8 & 1.0 &  1.8 &- & - & - &  65.6 &  5.7 &  10.4\\
& CONSE~\cite{norouzi2013zero} & 90.6 & 0.5 & 1.0 & 72.2 & 1.6 & 3.1 & 39.9 & 6.8 & 11.6 &- & - & - & 91.2 & 0.0 & 0.0\\
& CMT~\cite{socher2013zero}& 90 & 0.5 & 1 & 49.8 & 7.2 & 12.6 & 21.8 & 8.1 & 11.8 & - & - & - & 85.2 & 1.4 & 2.8 \\
& SSE~\cite{zhang2015zero}& 82.5 & 8.1 & 14.8 & 46.9 & 8.5 & 14.4 & 36.4 & 2.1 & 4 & - & - & - & 78.9 & 0.2 & 0.4\\
& ALE~\cite{akata2013label}& 81.8 & 14 & 23.9 & 62 & 23.7 & 34.4 & 33.1 & 21.8 & 26.3 & 61.6 & 13.3 & 21.9 & 73.7 & 4.6 & 8.7\\
& SAE~\cite{kodirov2017semantic}& 82.2 & 1.1 & 2.2 & 54.0 & 7.8 & 13.6 & 18.0  & 8.8 & 11.8 & - & - & - & 80.9 & 0.4 & 0.9\\
& EZSL~\cite{romera2015embarrassingly}& 77.8 & 5.9 & 11.0 & 63.8 & 12.6 & 21.0 & 27.9  & 11.0 & 15.8 & - & - & - & 70.1 & 2.4 & 4.6\\
& PSR~\cite{annadani2018preserving}& 73.8 & 20.7 & 32.3 & 54.3 & 24.6 & 33.9 & 37.2  & 20.8 & 26.7 & - & - & - & 51.4 & 13.5 & 21.4\\
\hline
\multirow{7}{*}{$\dagger$}&
f-CLSWGAN~\cite{xian2018feature}& 68.9 & 52.1 & 59.4 & 57.7 & 43.7 & 49.7 & 36.6 & 42.6 & 39.4 & 73.8 & 59.0 & 65.6 & 61.7 & 32.9 & 42.9\\
& Cycle-WGAN ~\cite{felix2018multi} & 63.4 & 59.6 & 59.8 & 59.3 & 47.9 & 53.0 & 33.8 & 47.2 & 39.4 & 69.2 & 61.6 & 65.2 & - & - & -\\
% & GAZSL~\cite{zhu2018generative} &  86.9 & 35.4 &  50.3 &  61.3 & 31.7 &  41.8 & 39.3 & 22.1 &  28.3 & - & - & - &  78.6 & 14.2 & 24.0\\
& SABR~\cite{paul2019semantically} & 93.9 & 30.3 & 46.9 & 58.7 & 55.0 & 56.8 & 35.1 & 50.7 & 41.5 & - & - & - & - & - & -\\
& f-VAEGAN-D2~\cite{xian2019f} &  70.6 & 57.6 &  63.5 & 60.1 & 48.4 & 53.6 & 38.0 & 45.1 & 41.3 & 74.9 & 56.8 & 64.6 & - & - & -\\
& LisGAN~\cite{li2019leveraging} & 76.3 & 52.6 & 62.3 & 57.9 & 46.5 & 51.6 & 37.8 & 42.9 & 40.2 & 83.8 & 57.7 & 68.3 & - & - & -\\
& Zero-VAE-GAN~\cite{gao2020zero} & 70.9 & 57.1 & 62.5 & 47.9 & 43.6 & 45.5 & 30.2 & 45.2 & 36.3 & - & - & - & 52.2 & 32.0 & 39.7\\
& TF-VAEGAN~\cite{narayan2020latent} & 75.1 & 59.8 & 66.6 & 64.7 & 52.8 & \textbf{\textcolor{blue}{58.1}} & 40.7 & 45.6 & 43.0 & 84.1 & 62.5 & \textbf{\textcolor{red}{71.7}} & 57.4 & 35.9 & \textbf{\textcolor{blue}{44.2}}\\
\hline
\multirow{6}{*}{$\ddagger$}& CVAE~\cite{mishra2018generative}& - & - & 51.2 & - & - & 34.5 & - & - & 26.7 & - & - & - & - & - & -  \\
& OCD-CVAE~\cite{keshari2020generalized}& 73.4 & 59.5 & 65.7 & 59.9 & 44.8 & 51.3 & 42.9 & 44.8 & \textbf{\textcolor{blue}{43.8}} & - & - & -  & - & - & -\\
& CADA-VAE~\cite{schonfeld2019generalized} &  75.0 & 55.8 & 63.9 & 53.5 & 51.6 & 52.4 & 35.7& 47.2 & 40.6 & 80.7 & 54.0 & 64.7 & 53.2 & 34.8 & 42.1 \\
& DE-VAE ~\cite{mavariational}& 78.9 & 58.8 & \textbf{\textcolor{red}{67.4}} & 56.3 & 52.5 & 54.3 & 36.9 & 45.9 & 40.9 & - & - & -  & - & - & -\\
\cline{2-17}
& Baseline & 75.0 & 55.8 & 63.9 & 57.3 & 49.7 & 53.3 & 35.5 & 49.0 & 41.2 & 77.1 & 54.2 & 63.7 &53.2 & 34.8 & 42.1\\
& DFS (Ours) & 78.6 & 58.4 & \textbf{\textcolor{blue}{67.2}} & 59.2 & 57.4 & \textbf{\textcolor{red}{58.3}} & 39.2 & 53.8 & \textbf{\textcolor{red}{45.4}} & 84.3 & 60.1 & \textbf{\textcolor{blue}{70.2}} & 60.7 & 37.1 & \textbf{\textcolor{red}{46.0}} \\
\hline
\end{tabular}
}
\end{table*}

\subsection{Comparison with SOTAs for GZSL}
We compare our model with recent state-of-the-art methods on generalized zero-shot learning, and the results are shown in Table~\ref{Tab:GZSLPer}. 
Since our whole model did not use any data from the testing dataset for training, only comparisons with inductive ZSL are made in all experiments.
For a fair comparison, we also trained a classifier using the features obtained from aligned feature generator and calculated the accuracy of its seen and unseen classes and the results can be seen on the model named Baseline.

Compared with baseline, significant improvement can be observed on all benchmarks.
The accuracy difference between our model and Baseline is as follows: \textbf{67.2}\% vs 63.9\% on AWA2, \textbf{58.3}\% vs 53.3\% on CUB, \textbf{45.4}\% vs 41.2\% on SUN, \textbf{70.2}\% vs 63.7\% on FLO and \textbf{46.0}\% vs 42.1\% on APY. 
At the same time, both $Acc_{s}$ and $Acc_{u}$ are improved, we attribute this high performance gain to the using of diverse samples for training classifier.
Compared with CVAE~\cite{mishra2018generative}, we use the aligned space as classification space while using the aligned semantic information as the condition.
These changes resulted in  16\%, 23.8\% and 18.7\% improvement in our network over CVAE for AWA2, CUB and SUN, respectively.
The model f-VAEGAN-D2~\cite{xian2019f} which based on VAE-GAN reports classification accuracies of 63.5\%, 53.6\%, 41.3\% and 64.6\% on AWA2, CUB, SUN, and FLO, respectively.
The improved model of f-VAEGAN-D2, TF-VAEGAN~\cite{narayan2020latent}, obtains state-of-the-art classification scores of 66.6\%, 58.1\%, 43.0\% and 71.7\% on the same datasets. 
But it is worth noting that it introduces GAN and applies more complex training process, which often makes the model unable to learn stably.
In contrast, all of our modules are built on VAE. Hence, our method is simple and can be trained stably without using any training skills.

In addition, DFS outperforms TF-VAEGAN 0.6\%, 0.2\%, 2.4\% and 1.8\% on four datasets and set new state-of-the-art.
Similarly to our work, DE-VAE~\cite{mavariational} is also an improved model based on CADA-VAE.
However, except for AWA2, the performance of our model is far better than it.
Since the motivation of DFS and DE-VAE do not conflict with each other, we speculate that integrating the two methods together will yield superior results.
Nevertheless, this experiment cannot be performed in this paper because the code of DE-VAE is not available.
In the conventional zero-shot learning, DFS also provides favourable performance, 69.1\% on AWA2, 64.7\% on CUB, 64.4\% SUN, 68.9\% on FLO and 43.6\% on APY.
Nevertheless, we focus on the more practical and challenging GZSL setting in this work.

\subsection{Ablation Study}
\begin{figure}[t!]
\begin{center}
  \includegraphics[width=0.45\textwidth]{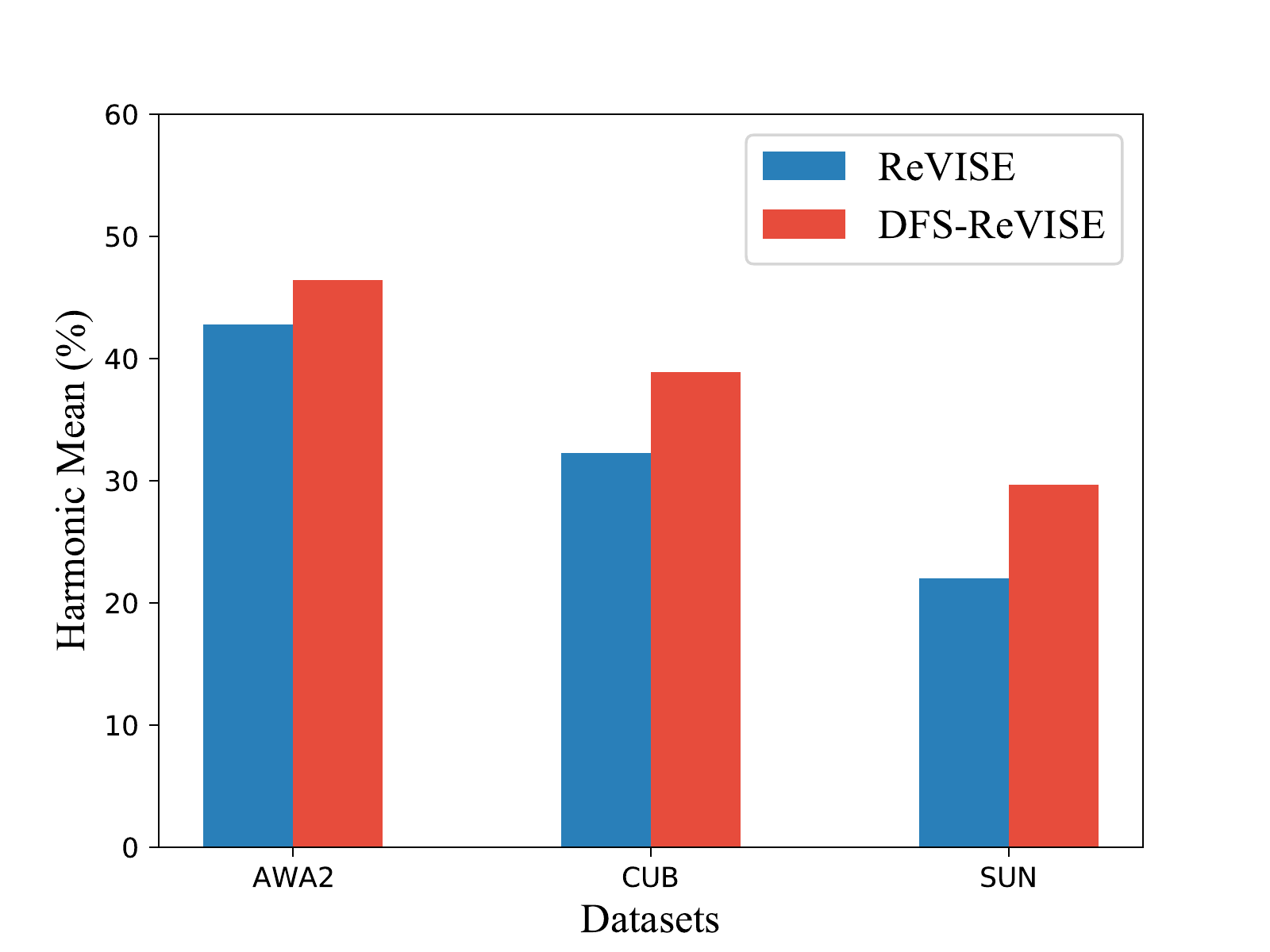}
\end{center}
  \caption{Verification on the generalization capability of our DFS model. Here, we replace CADA-VAE with ReVISE to generate the aligned features and exploit the harmonic mean accuracy ($\%$) as the metric. Results on AWA2, CUB and SUN datasets are shown. (Best viewed in color)
  %Instead of using CADA-VAE to learn aligned space, we integrate our proposed contributions with ReVISE and then measure the harmonic mean accuracy (\%) on the AWA2, CUB and SUN. (Best viewed in colour)
  }
\label{fig:generate_ability}
\end{figure}

\paragraph{Generalization Capabilities}As mentioned in Section~\ref{sec:first_stage}, our innovation point is not limited to a specific cross model, but can be directly introduced into most of the previously proposed cross models.
To verify the generalizability of our idea, we perform an experiment by integrating the contributions proposed in this work in ReVISE~\cite{hubert2017learning}.
The results in Figure~\ref{fig:generate_ability} show that DFS-ReVISE outperforms ReVISE on all three datasets.
These performance gains entirely benefit from the fact that we synthetise more diverse samples for unseen classes in latent space, which is helpful to learn an effective classifier.
Since CADA-VAE is the powerful model to learn aligned space in GZSL, we implement AFG with CADA-VAE in the following experiments for discussion.

\begin{figure}[ht]
\begin{center}
  \includegraphics[width=0.45\textwidth]{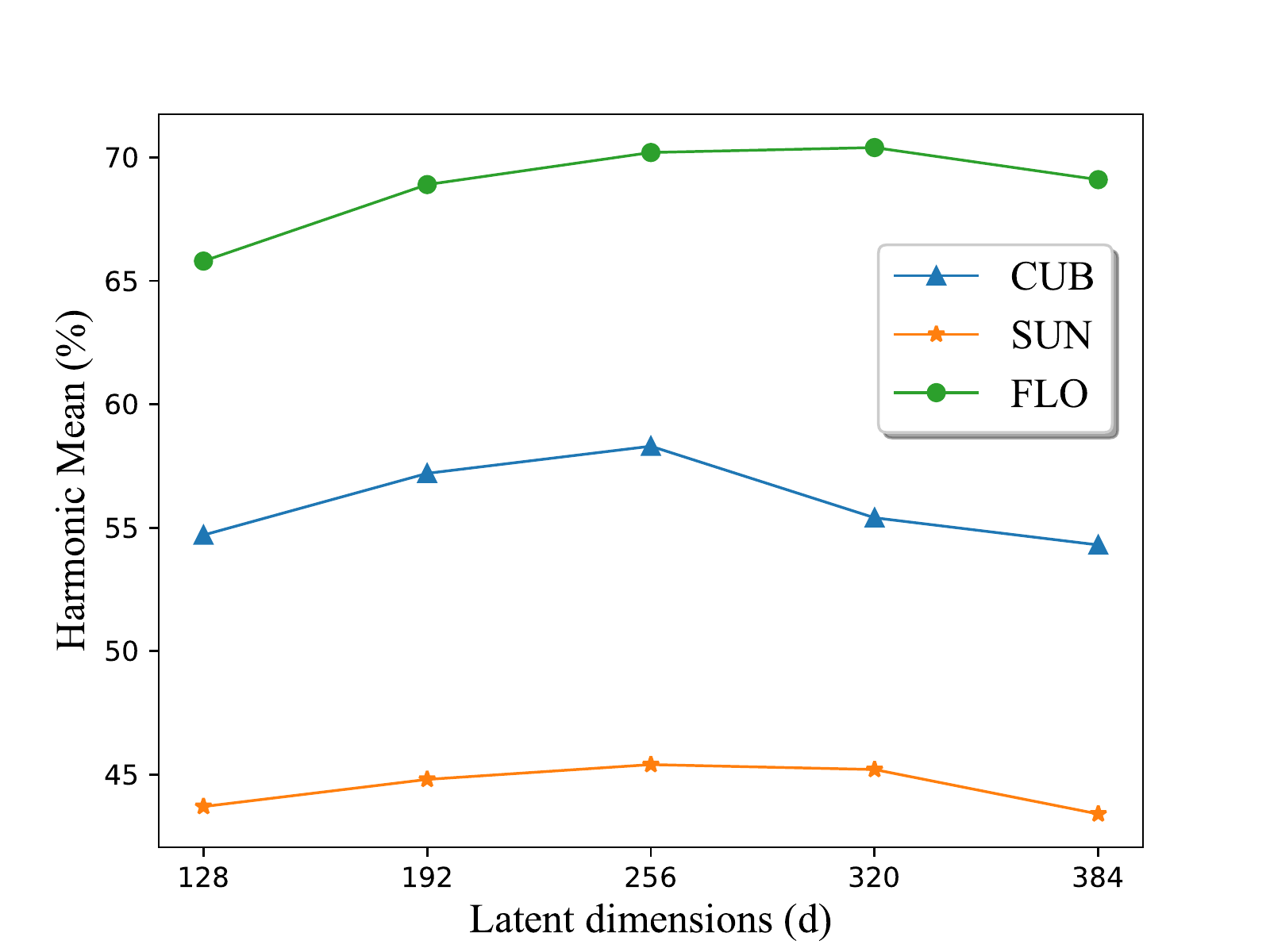}
\end{center}
\vspace{-5mm}
  \caption{Ablation study on the effect of the dimentionality of the aligned features on CUB, SUN and FLO datasets. Harmonic mean accuracy is used as the metric.
  %The reported values are harmonic mean accuracy (\%). 
  (Best viewed in color)}
\label{fig:diff_dim}
  \vspace{-5mm}
\end{figure}

\paragraph{Dimension of Aligned Space} Figure~\ref{fig:diff_dim} presents the summary statistics of accuracy under different dimensionality of the aligned space on three datasets.
It can be observed that with the increasing dimensionality, the accuracy of DFS increases initially until $d=256$, $d =256$ and $d =320$ for CUB, SUN and FLO, respectively.
Intuitively, higher dimensions tend to contain more complex information, such as information specific to visual space.
It is our assertion that part of the visual-specific information is beneficial for the classification task because similar categories may not be classified by domain shared information.
The visual space features usually come from the powerful model trained on large-scale datasets and thus will contain more information that can effectively distinguish between different categories.
Although these information is not directly derived from the semantic features, we believe that some of it can be interpreted by semantic features. 
In our work, SFG aids to capture the distribution of visual-specific information that can be reasonably inferred by class embeddings.
However, the distribution will become extremely complex while the number of dimension is too large, which will result in SFG unable to learn the real distribution, so the performance begins to decline.
In order to make the model not lose generality, we set the $d=256$ for all fine-grained datasets.

\paragraph{Feature Visualization}
\begin{figure*}[ht!]
\subcaptionbox{\label{fig:seen_classes}}[0.33\textwidth]
    { 
        \includegraphics[width=0.33\textwidth]{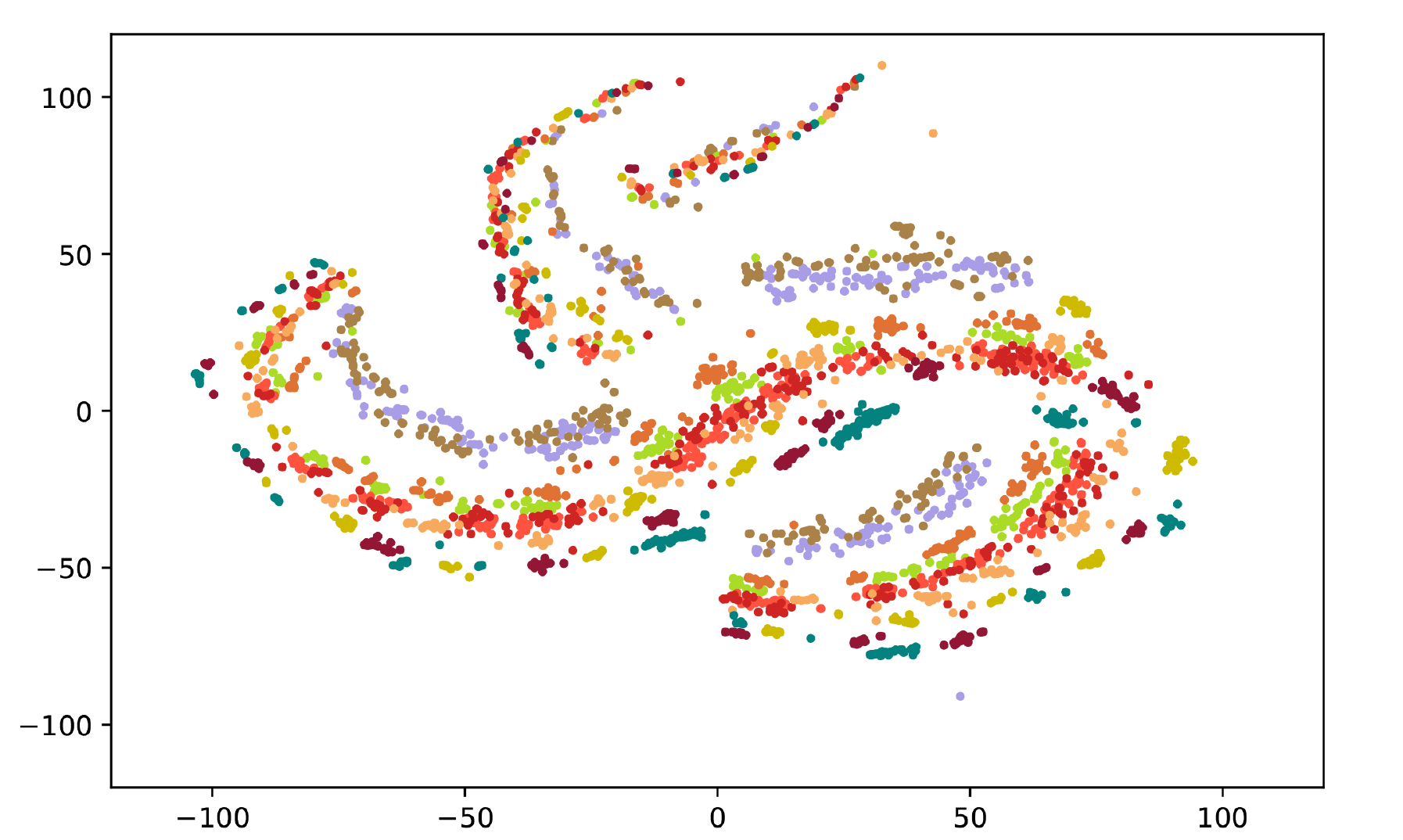}
        \includegraphics[width=0.33\textwidth]{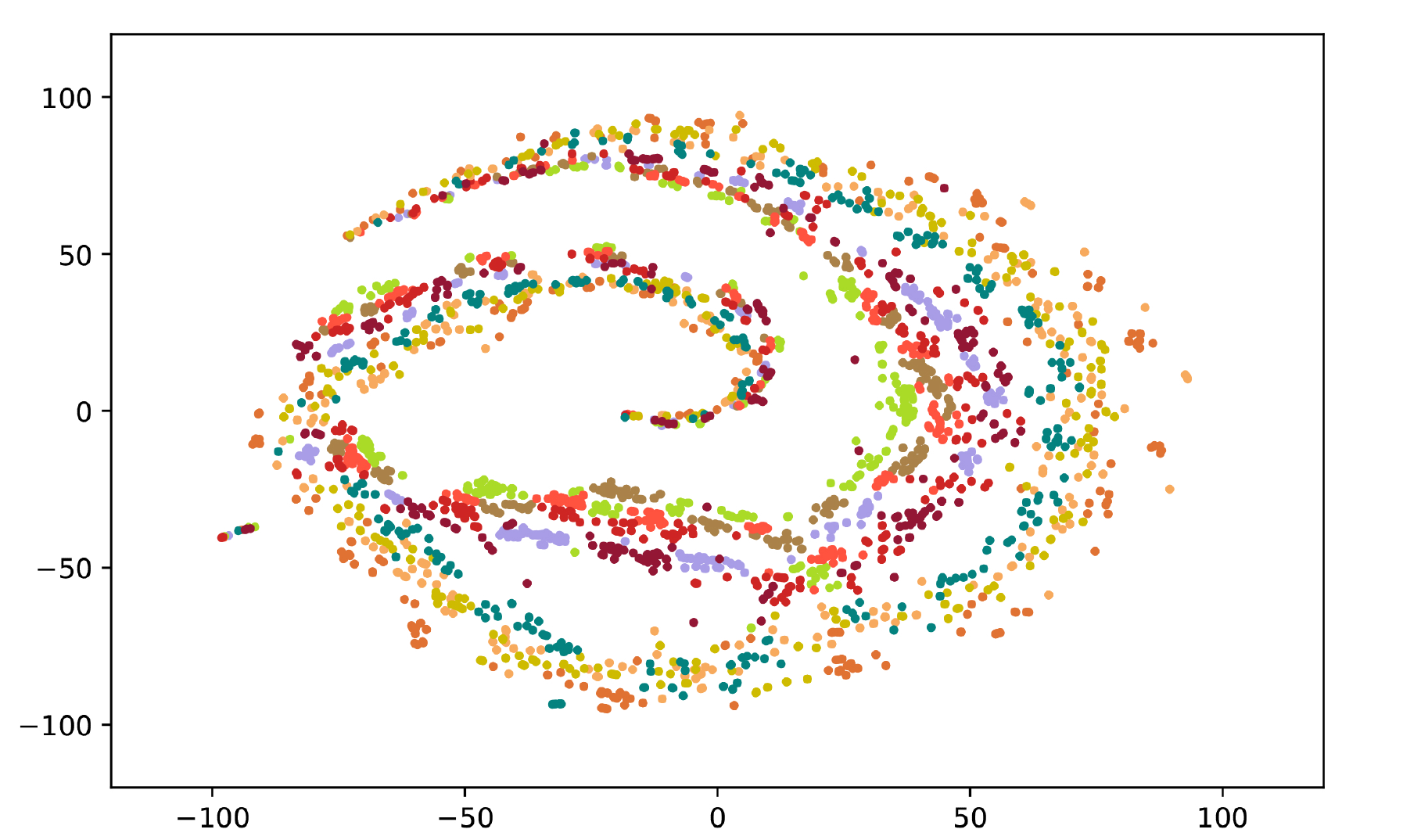}
        \includegraphics[width=0.33\textwidth]{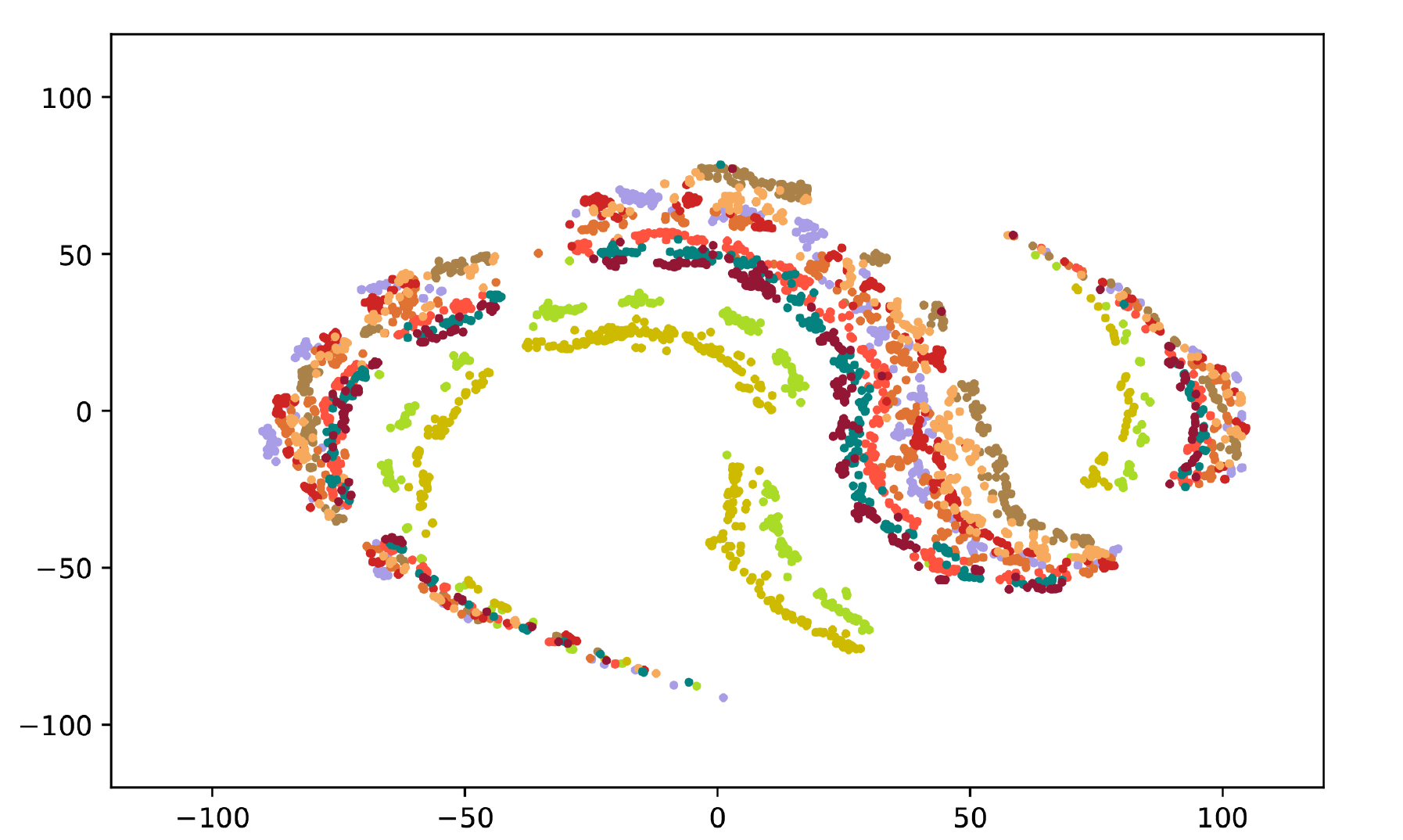}
    } 
\subcaptionbox{\label{fig:unseen_classes_ori}}[0.33\textwidth]
    { 
        \includegraphics[width=0.33\textwidth]{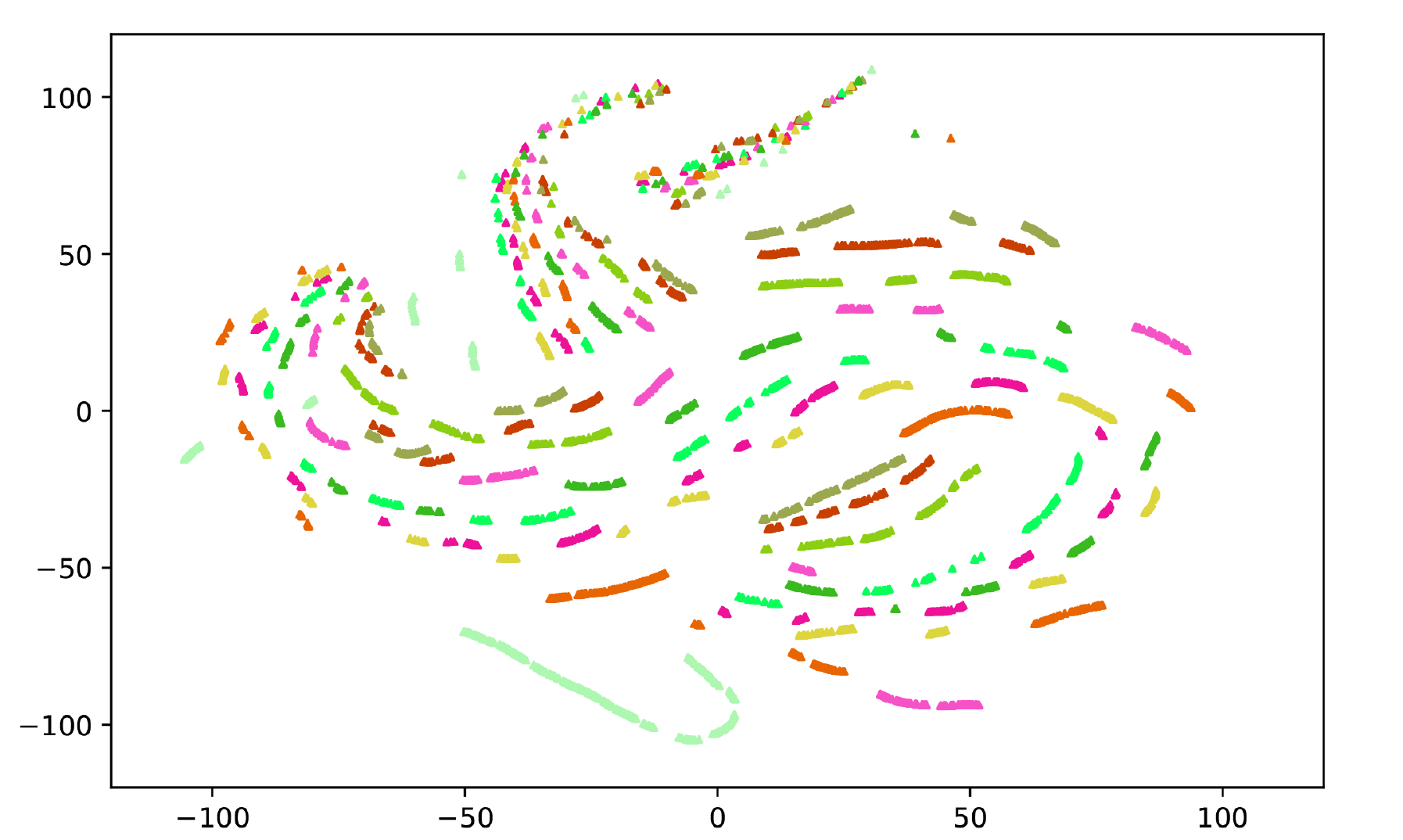}
        \includegraphics[width=0.33\textwidth]{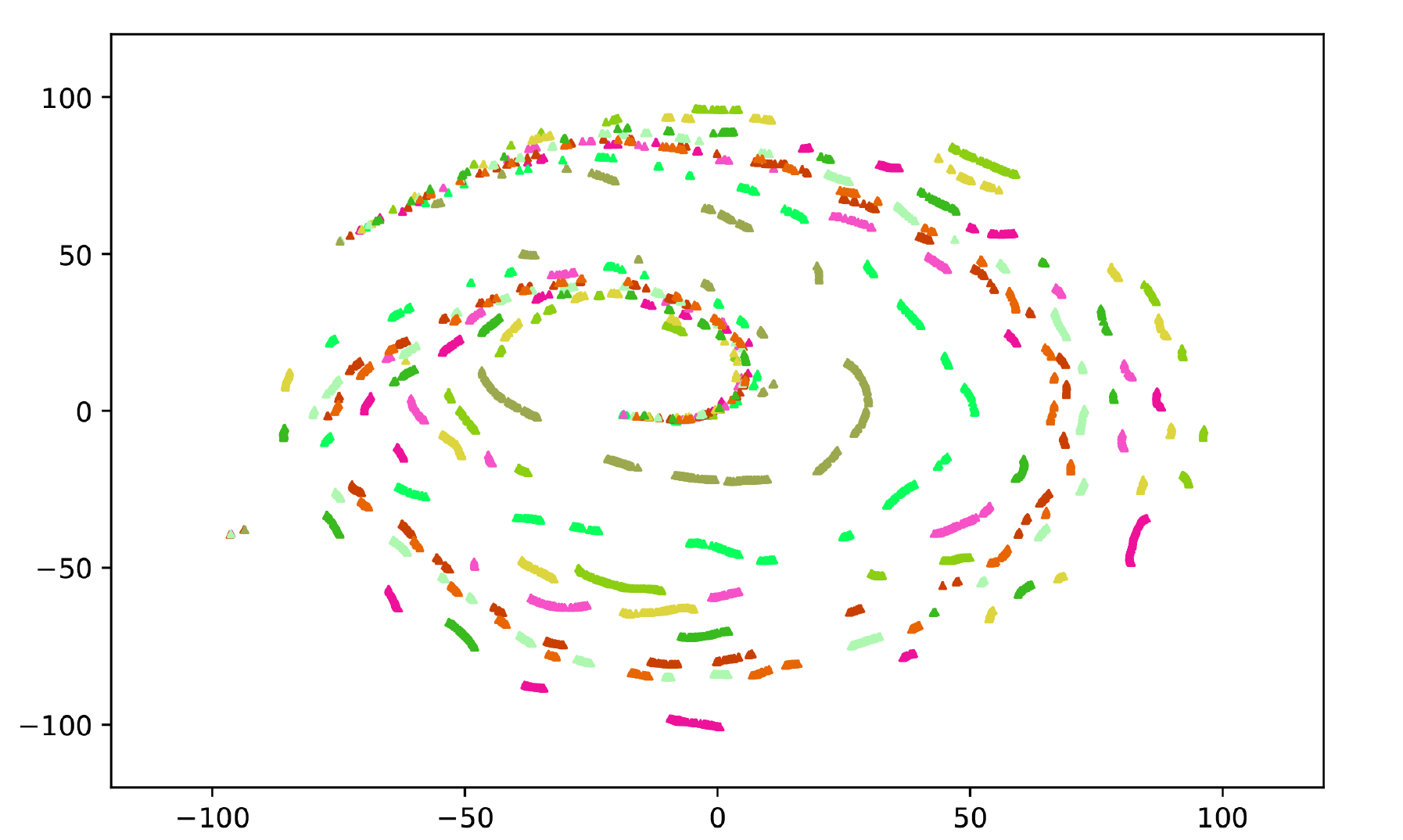}
        \includegraphics[width=0.33\textwidth]{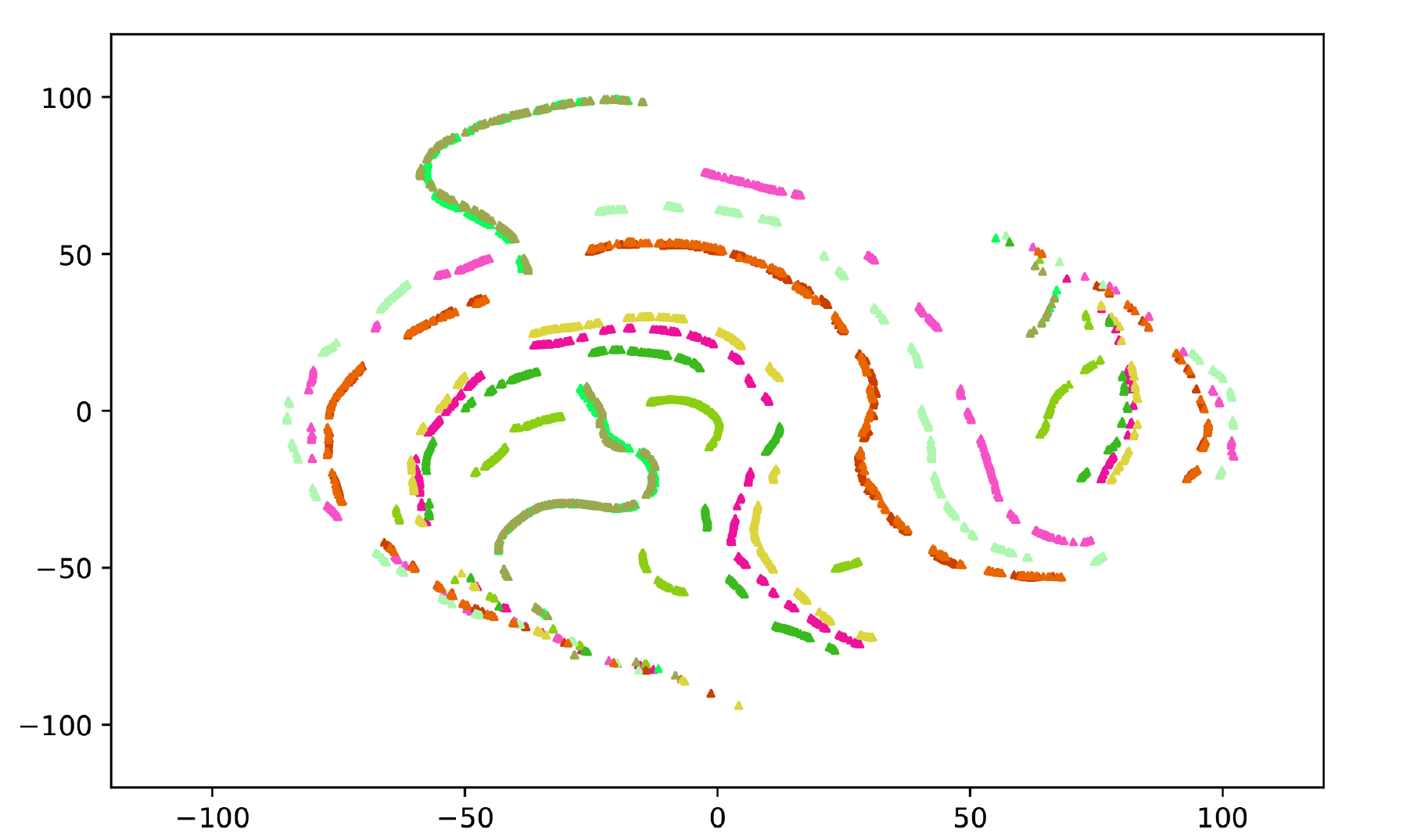}
        
    }
\subcaptionbox{\label{fig:unseen_classes_our}}[0.33\textwidth]
    { 
        \includegraphics[width=0.33\textwidth]{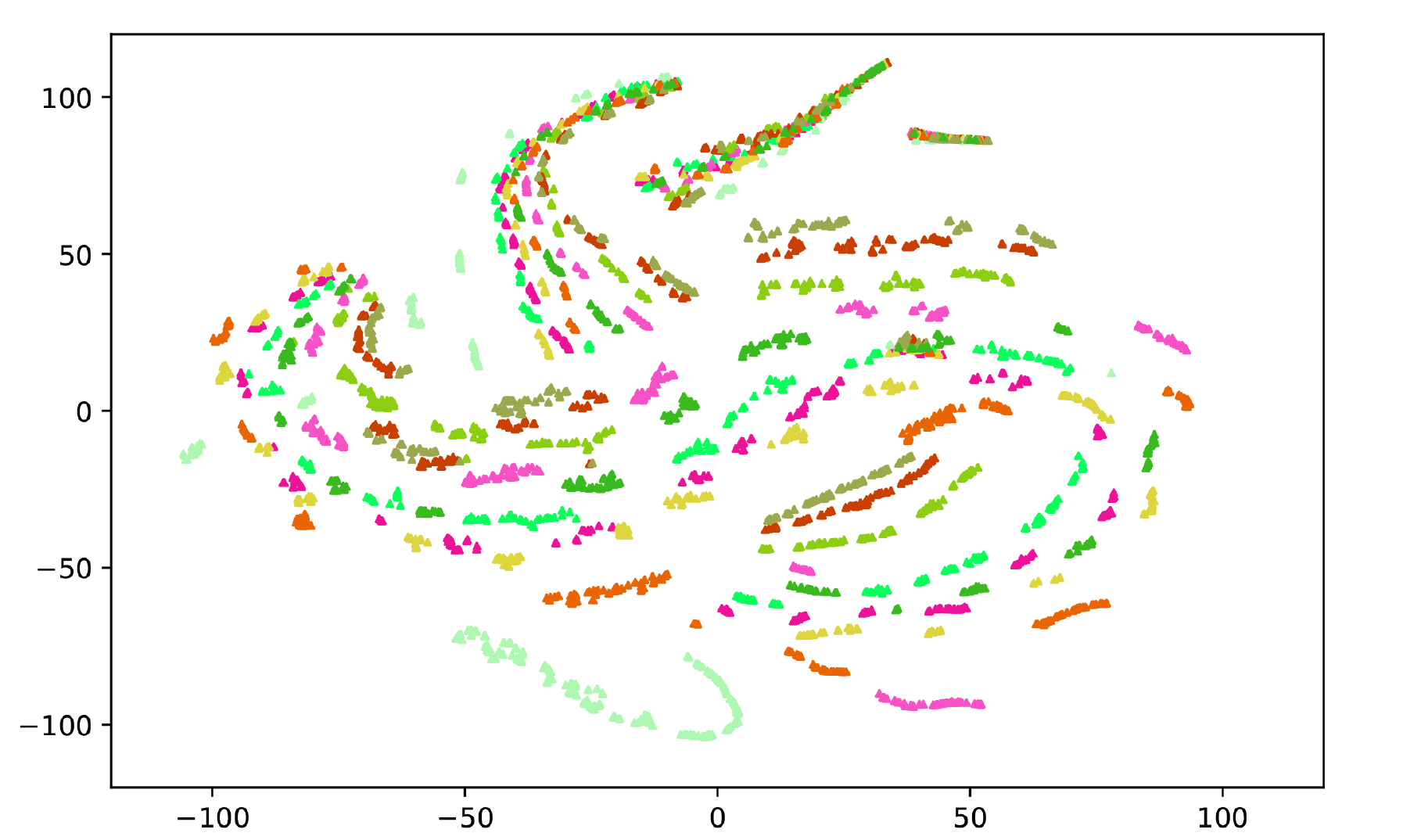}
        \includegraphics[width=0.33\textwidth]{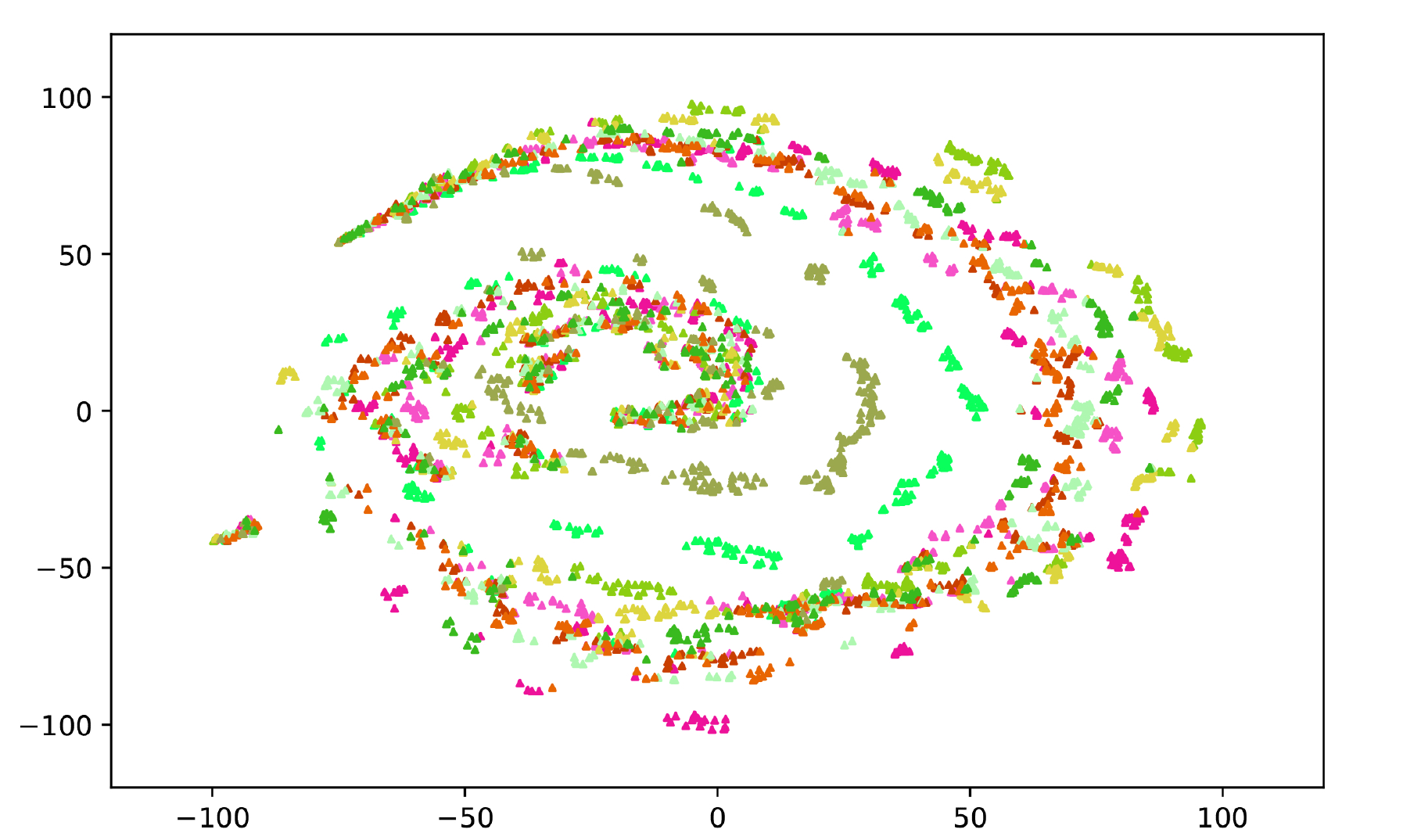}
        \includegraphics[width=0.33\textwidth]{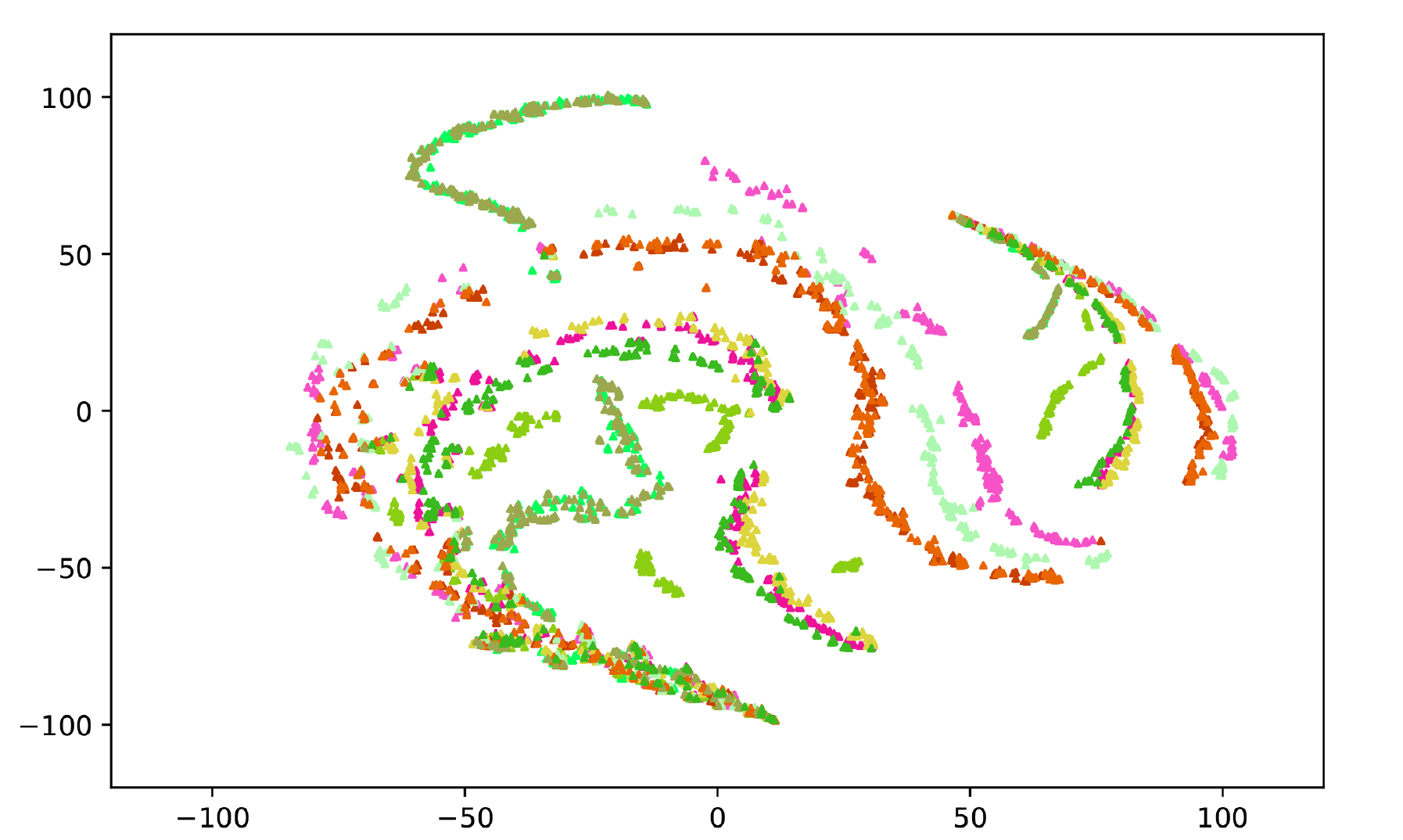}
    }
\caption{Visualization of samples in aligned space with t-SNE on CUB (1st row), SUN (2nd row) and APY (3rd row). (a) Samples generated for seen classes. (b) Samples generated for unseen classes by Baseline. (c) Samples generated for unseen classes by DFS. (Best viewed in color)}
\end{figure*}
To verify that the samples generated by DFS for unseen classes are more diverse, we visualized the features in the aligned space by using t-SNE~\cite{van2008visualizing}.
For each dataset, we randomly selected 10 seen classes and 10 unseen classes, respectively.
Each seen class takes 400 samples in aligned space using multiple visual features.
And for each unseen class, 400 samples are generated by semantic vectors.
It is not difficult to find by Figure~\ref{fig:seen_classes} and Figure~\ref{fig:unseen_classes_ori}, compared to the seen classes, the diversity of the unseen classes instances generated by Baseline is far from adequate.
When the two classes are closer to each other, the classifier would tend to predict the test sample as the class with greater diversity.
Figure~\ref{fig:unseen_classes_ori} and Figure~\ref{fig:unseen_classes_our} show a comparison between Baseline and our methods.
Obviously, in all three datasets, the instances synthesised by DFS are more diverse.
This will help the classifier learn the decision boundary with better generalization performance.

\begin{table}[ht]
\caption{Ablation study on the influence of different features as conditions on performance. $a$ denotes the original semantic feature, $\mu^{1}$ and $\sigma^{1}$ are the parameters computed by $E^{1}$ with $a$ as input. Harmonic mean accuracy is used as the metric.}
\label{Tab:diff_condition}
\centering
\begin{tabular}{c|c|c|c}
\hline
Condition & CUB & FLO & APY\\
\hline
$a$ & 56.4 & 68.6 & 40.9 \\
$z^{1}\sim \mathcal{N}(\mu^{1},\sigma^{1})$ & 55.0 & 69.2 & 42.7\\
$z^{1} = \mu^{1}$ & \textbf{58.3} & \textbf{70.2} & \textbf{46.0}\\
\hline
\end{tabular}
\end{table}

\paragraph{Choice of Condition} 
To further justify the influence of different supervision signals on the performance of the model, we use different features as condition to train SFG module. 
It is apparent from Table~\ref{Tab:diff_condition} that using the mean of semantic features in aligned space as supervised information provides the best results on all three datasets.
We analyze that this is mainly attributed to the fact that $z^{1}$ is on the same manifold as the target distribution to be learned, which significantly reduces the training difficulty of the generator.
But at the same time, if we let $z^{1}\sim \mathcal{N}(\mu^{1},\sigma^{1})$, it will make an unstable condition and therefore interfere with the learning process of the generator.

\section{Conclusion}
\label{con}
In this paper, we present a novel Diverse Feature Synthesis (DFS) model for enhancing the generalizability of generative based strategy for the generalized zero-shot learning task. In particular, DFS effectively improves the feature diversity of unseen classes with a low-complexity implementation. For this purpose, DFS first utilizes an aligned feature generator to transfer features from semantic and visual spaces into the aligned space, offering a way for simplifying the feature generation process. Then, DFS exploits a synthesised feature generator to produce aligned features for unseen classes via incorporating visual knowledge, thus leading to feature diversity improvement. In this way, DFS overcomes drawbacks of prior works, and helps to learn a more accurate and robust classifier for the GZSL task. Comprehensive experiments on multiple benchmarks verify the effectiveness of our proposed DFS model for improving the performance in GZSL settings. 

% {\small
% \bibliographystyle{ieee}
% \bibliography{egbib}
% }

\end{document}